\documentclass{article}

\usepackage[preprint]{neurips_2025}
\usepackage{multirow}
\usepackage{graphicx}
\usepackage{svg}
\usepackage{amsmath}
\usepackage{caption} 
\usepackage{algorithm}
\usepackage{listings}
\usepackage[utf8]{inputenc} 
\usepackage[T1]{fontenc}    
\usepackage{hyperref}       
\usepackage{url}            
\usepackage{booktabs}       
\usepackage{amsfonts}       
\usepackage{nicefrac}       
\usepackage{microtype}      
\pdfoutput=1

\title{
MonoVQD: Monocular 3D Object Detection with Variational Query Denoising and Self-Distillation
}

\author{%
  Kiet Dang Vu, Trung Thai Tran, and Duc Dung Nguyen\thanks{Corresponding}\\
  AITech Lab, Computer Science and Engineering Faculty\\
  Ho Chi Minh City University of Technology, VNUHCM\\
  \texttt{\{kiet.dangvutuan0712, thai.tran241002, nddung\}@hcmut.edu.vn}\\
}

\begin{document}

\maketitle

\begin{abstract}

  Precisely localizing 3D objects from a single image constitutes a central challenge in monocular 3D detection. While DETR-like architectures offer a powerful paradigm, their direct application in this domain encounters inherent limitations, preventing optimal performance. Our work addresses these challenges by introducing MonoVQD, a novel framework designed to fundamentally advance DETR-based monocular 3D detection. We propose three main contributions. 
  First, we propose the Mask Separated Self-Attention mechanism that enables the integration of the denoising process into a DETR architecture. This improves the stability of Hungarian matching to achieve a consistent optimization objective. Second, we present the Variational Query Denoising technique to address the gradient vanishing problem of conventional denoising methods, which severely restricts the efficiency of the denoising process. This explicitly introduces stochastic properties to mitigate this fundamental limitation and unlock substantial performance gains.
  Finally, we introduce a sophisticated self-distillation strategy, leveraging insights from later decoder layers to synergistically improve query quality in earlier layers, thereby amplifying the iterative refinement process.
   Rigorous experimentation demonstrates that MonoVQD achieves superior performance on the challenging KITTI monocular benchmark. Highlighting its broad applicability, MonoVQD's core components seamlessly integrate into other architectures, delivering significant performance gains even in multi-view 3D detection scenarios on the nuScenes dataset and underscoring its robust generalization capabilities.
\end{abstract}

\section{Introduction}
Accurate 3D object detection stands as a cornerstone of autonomous driving systems, providing the essential capability to precisely perceive and understand the surrounding environment. This understanding, encompassing the precise localization, dimensional attributes, and spatial orientation of crucial objects such as vehicles and pedestrians, is paramount for ensuring safe navigation and enabling well-informed decision-making processes within autonomous vehicles. While methodologies leveraging the high-fidelity depth information offered by LiDAR sensors~\cite{PeP,Centerbased3D,PV-RCNN} and sophisticated multi-camera configurations~\cite{RayDN,BEVFormer,PETR} have showcased superior performance in this domain, these multi-sensor approaches inherently present certain limitations. Their dependence on multiple sensing modalities introduces increased system complexity, elevates the potential for sensor failures or miscalibration issues, and consequently can impede their widespread deployment, particularly in cost-sensitive application scenarios. Therefore, monocular 3D object detection emerges as a highly compelling, inherently robust, and practically advantageous alternative for resource-constrained deployments, as it necessitates the utilization of only a single, ubiquitous camera sensor.

\begin{figure}[t!]
    \centering
    \includegraphics[width=\linewidth]{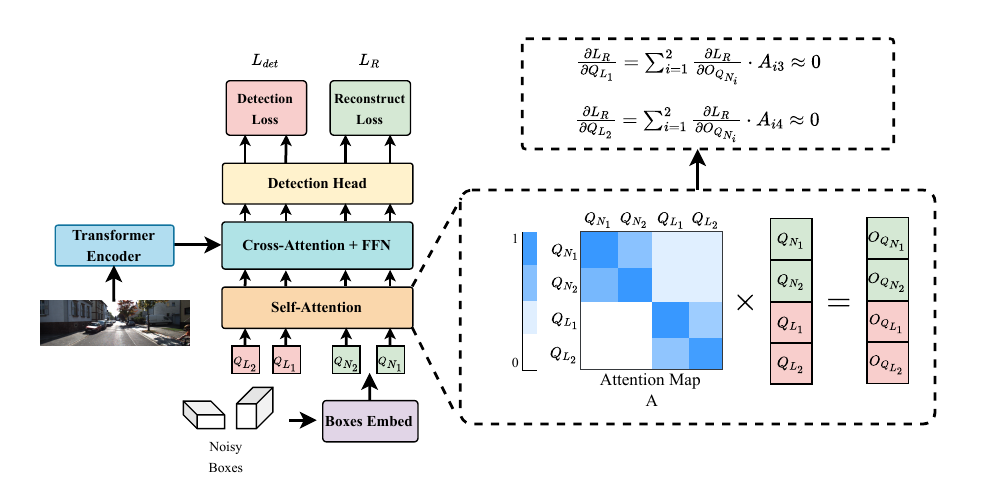}
    \caption{\textbf{Limitations of conventional denoising approach.} During the later stages of training, the attention score between noisy queries and learnable queries diminishes to zero. This causes the gradient from the reconstruction loss with respect to the learnable queries to also become zero.}
    \label{fig:intro}
\end{figure}

Despite advancements in monocular 3D object detection~\cite{DCD,DID-M3D,MonoCD,MonoUNI,MonoJSG,MonoDDE}, the inherent lack of direct depth information from single-view images remains a significant challenge. To mitigate this limitation, several studies have focused on incorporating estimated depth maps to guide the detection learning process~\cite{MonoDTR,MonoDETR,MonoPGC,FD3D}. Notably, MonoDETR~\cite{MonoDETR} introduced a DETR-based framework for monocular 3D detection, employing a depth-guided transformer architecture. While this approach demonstrated improved object localization compared to previous methods, there is still room to improve the performance of DETR framework for monocular 3D object detection.

Inspired by the denoising training principle introduced in DN-DETR~\cite{DN-DETR}, which effectively enhances object detection performance by leveraging noisy ground truth bounding boxes for decoder reconstruction, we extended this approach to the domain of 3D detection within MonoDETR. This extension was realized by generating noisy 3D bounding boxes from the ground truth annotations and incorporating them as input to the decoder. However, this adaptation introduced a significant challenge stemming from the base DETR architecture's reliance on the Group-DETR strategy~\cite{GroupDETR}, which is crucial for achieving rapid convergence. To successfully integrate the performance benefits of denoising with the training efficiency provided by group-based attention, we designed the Mask Separated Self-Attention mechanism. Furthermore, as illustrated in Fig.~\ref{fig:intro}, a gradient vanishing problem occurs, affecting the gradient flow from the reconstruction loss to the learnable queries, thereby limiting the impact of the denoising process on the model. To mitigate this limitation and secure additional performance gains, we proposed the Variational Query Denoising. Moreover, we introduced a novel Forward-Looking Distillation method aimed at improving the iterative refinement capability of the base DETR model~\cite{DETR} in the early decoder layers. Importantly, MonoVQD incurs no additional computational cost or parameters during inference, thus maintaining efficiency critical for real-world deployment.

This work presents four primary contributions. \textit{First,} we propose a novel Mask Separated Self-Attention synergistically combining multi-group query and denoising paradigms, providing fine-grained control over noisy/learnable query interaction during group aggregation. \textit{Second,} we introduce Variational Query Denoising, a training scheme addressing gradient vanishing in denoising by encoding noisy 3D boxes as stochastic latent queries for stable training and improved accuracy. \textit{Third,} We propose Forward-Looking Distillation, a novel self-distillation method enhancing early decoder query quality using later layers, further improving DETR's iterative refinement. 

Extensive experimental results establish MonoVQD as the new state-of-the-art on the challenging KITTI monocular benchmark, significantly advancing the field. We further demonstrate its broad applicability and modular integrability by showing substantial performance gains when integrated into multi-view architectures on the nuScenes dataset, highlighting exceptional generalization.

\section{Related Work}
\textbf{Detection Transformer.} Since the introduction of the Detection Transformer (DETR)~\cite{DETR}, significant progress has been made in object detection. Subsequent work has addressed its limitations, such as slow convergence and limited spatial resolution, through various innovations. Deformable DETR~\cite{DeformableDETR} replaced the original attention mechanism with deformable attention for more efficient feature sampling. DN-DETR~\cite{DN-DETR} introduced a denoising training scheme to stabilize the Hungarian matching from inconsistent optimization goals during early training. GroupDETR~\cite{GroupDETR} further improved training stability and performance by incorporating one-to-many matching methods, providing additional positive supervision. In this work, we extend these advancements in DETR-based methods to monocular 3D object detection, with a particular focus on enhancing the denoising strategy.

\textbf{Monocular 3D Object Detection.} Inferring object depth from single 2D images is a challenge for monocular 3D detection. Researchers explore diverse methods to mitigate this. Initial approaches~\cite{monoflex,DCD,MonoCD} enhanced depth by generating multiple candidates from geometry-inspired techniques and applying depth ensembling for a refined value. More recent efforts~\cite{MonoDTR,MonoDETR,MonoPGC,FD3D} integrate transformer architectures. These methods typically extract visual and depth features using a backbone and lightweight predictor, which are then processed by transformer encoders and aggregated in the decoder for robust detection. MonoDETR~\cite{MonoDETR} is an initial application of the DETR framework for monocular 3D object detection. It predicts foreground depth leveraging object-level annotations, integrates a depth-guided decoder, and uses object queries for effective global feature aggregation. However, MonoDETR's performance can be improved. This research introduces an enhanced training approach boosting base DETR detection capabilities without increased inference overhead.

\textbf{Multi-view 3D Object Detection.} DETR3D~\cite{DETR3D} was proposed for the joint extraction of features from surrounding views. Extending this line of work, the PETR series~\cite{PETR,liu2023petrv2} further introduced the generation of 3D positional features without relying on unstable projection operations and explored the benefits of incorporating temporal information from preceding frames. Nevertheless, these methods struggle with precise prediction due to depth ambiguities, often causing false positives along camera rays. RayDN~\cite{RayDN} addresses this via depth-aware hard negative sampling along rays to learn more discriminative features. In contrast to RayDN \cite{RayDN}, MonoVQD is specifically designed for monocular images. Nonetheless, our proposed method exhibits generalizability and can be integrated in a plug-and-play manner to enhance multi-view detectors.
\begin{figure*}[t!]
    \centering
    \includegraphics[width=\linewidth]{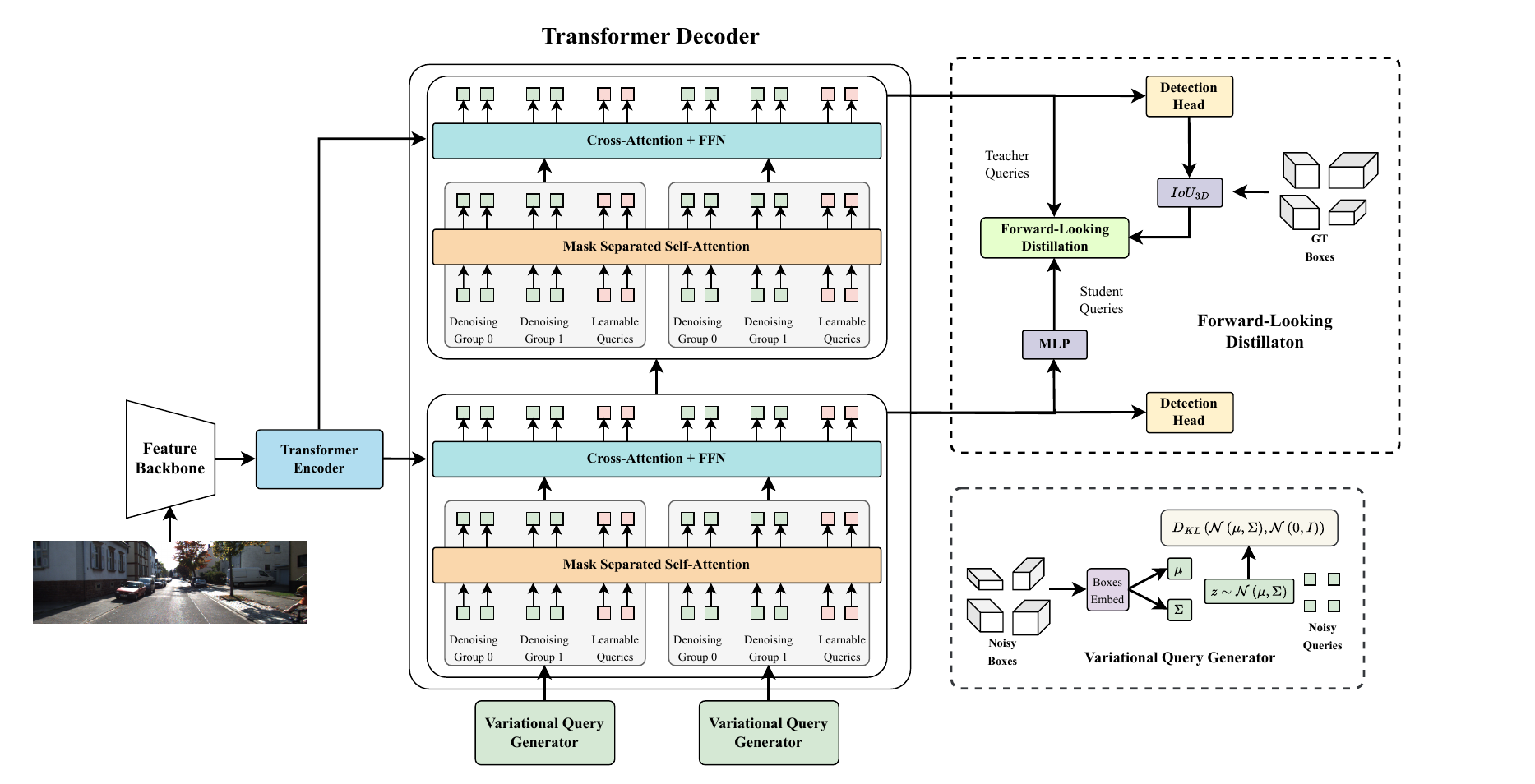}
    \caption{\textbf{The overall of our proposed framework MonoVQD.} Employing a pretrained image backbone to extract initial feature representations, the architecture subsequently processes these features through a transformer encoder. The decoder then operates on groups of trainable query features, with each group having corresponding noisy query features generated by a Variational Query Generator. These trainable and noisy queries are processed by a Mask Separated Self-Attention mechanism before undergoing cross-attention with the encoded features. Finally, Forward-Looking Distillation enhances early decoder performance by leveraging guidance from later layers.}
    \label{fig:overall}
    \vspace*{-\baselineskip}
\end{figure*}
\section{Methodology}
\label{sec:method}
\subsection{Overview}
\label{sec:overview}
Fig.~\ref{fig:overall} illustrates the architecture of our proposed framework. Given a single-view image $I \in \mathbb{R}^{H \times W \times 3}$, we first employ ResNet50~\cite{ResNet} to extract high-level feature maps, which are subsequently input to a transformer encoder. Following the baseline MonoDETR\cite{MonoDETR}, the decoder utilizes $G$ groups of learnable queries $Q_L = \{q_{L_i}\}_{i=1}^G$, where $q_{L_i}\in\mathbb{R}^{N \times D}$, with $N$ representing the number of queries in each group and $D$ denoting the hidden dimension of these queries. For each of these $G$ groups of learnable queries, we introduce $C$ corresponding groups of noisy queries $Q_N = \{\{q_{N_{ij}}\}_{j=1}^C\}_{i=1}^G$, where for the $i$-th group of learnable queries, we have $C$ associated groups of noisy queries, and $q_{N_{ij}}\in\mathbb{R}^{K \times D}$, with $K$ signifying the number of objects present in the input image. These noisy queries are generated by a Variational Query Generator. Both the learnable queries $Q_L$ and their corresponding noisy queries $Q_N$ are processed by a Mask Separated Self-Attention mechanism to model the interactions within each set of queries. Subsequently, a cross-attention module aggregates information from the transformer encoder. To enhance the performance of the initial decoder layers, we introduce Forward-Looking Distillation, where the later layers provide guidance, further refining the iterative refinement process inherent in the DETR framework. It is important to note that during inference, the noisy queries $q_N$ are not utilized.
\subsection{3D Query Denoising}
\label{sec:3D_DENOISING}
\textbf{3D Noisy Query.} In 2D object detection, DN-DETR~\cite{DN-DETR} utilizes the bounding box as the reference anchor and category of object to generate the noise query. We first reformulate how to generate a noisy query to enhance the denoising technique for 3D object detection. The objective of monocular 3D object detection consist of: category $c$, projected center $\left(x_c,y_c\right)$, 2D bounding box $l,r,t,b$, 3D dimension $l_{3D},w_{3D},h_{3D}$, orientation $\theta$ and central depth $d$. After generating noise boxes, we denote a 6D anchor box $\left(x_c,y_c,l,r,t,b\right)$ as the initial reference. Then we map the 3D information $\left(c,l_{3D},w_{3D},h_{3D},\theta,d\right)$ to a hidden space using an embedding layer, yielding the 3D query. The anchor box and the query are fed into the decoder to be reconstructed. The specific details of the embedding layer and the noise generation mechanism are provided in the supplementary material.

\textbf{Mask Separated Self-Attention.} To improve detection accuracy, the baseline MonoDETR model incorporates a group-wise one-to-many assignment strategy~\cite{GroupDETR}, employing $G$ distinct groups of learnable queries $Q_L = \{q_{L_i}\}_{i=1}^G$. This approach utilizes Separate Self-Attention~\cite{GroupDETR} to ensure that queries belonging to different groups do not interact, thereby maintaining group independence. To leverage the benefits of both the denoising training paradigm and the group-wise assignment methodology, we introduce the Mask Separated Self-Attention mechanism. This mechanism employs a predefined attention $M \in \{0,1\}^{S \times S}$ inspired by DN-DETR~\cite{DN-DETR}, where $S = K \cdot C + N$, to regulate the interactions only among noisy queries, as well as between noisy queries and learnable queries. Specifically, from the sets of learnable queries $Q_L$ and noisy queries $Q_N$, we construct a new set of combined queries $Q = \{q_i\}_{i=1}^G$, where each $q_i \in \mathbb{R}^{S \times D}$ is constructed by concatenating the learnable query $q_{L_i}$ with its corresponding noisy queries $\{q_{N_{ij}}\}_{j=1}^C$:
\begin{equation}
    q_i = \text{concat}(q_{L_i}, q_{N_{i1}}, q_{N_{i2}}, ..., q_{N_{iC}})
    \label{eq:q_i}
\end{equation}
This newly formed set of queries is then processed by Separated Self-Attention with the predefined mask $M$. Through this proposed Mask Separated Self-Attention, we achieve precise control over the interaction patterns between queries of different types and within each group of queries.
\subsection{Variational Query Denoising}
\begin{figure*}[t!]
    \centering
    \includegraphics[width=0.9\linewidth]{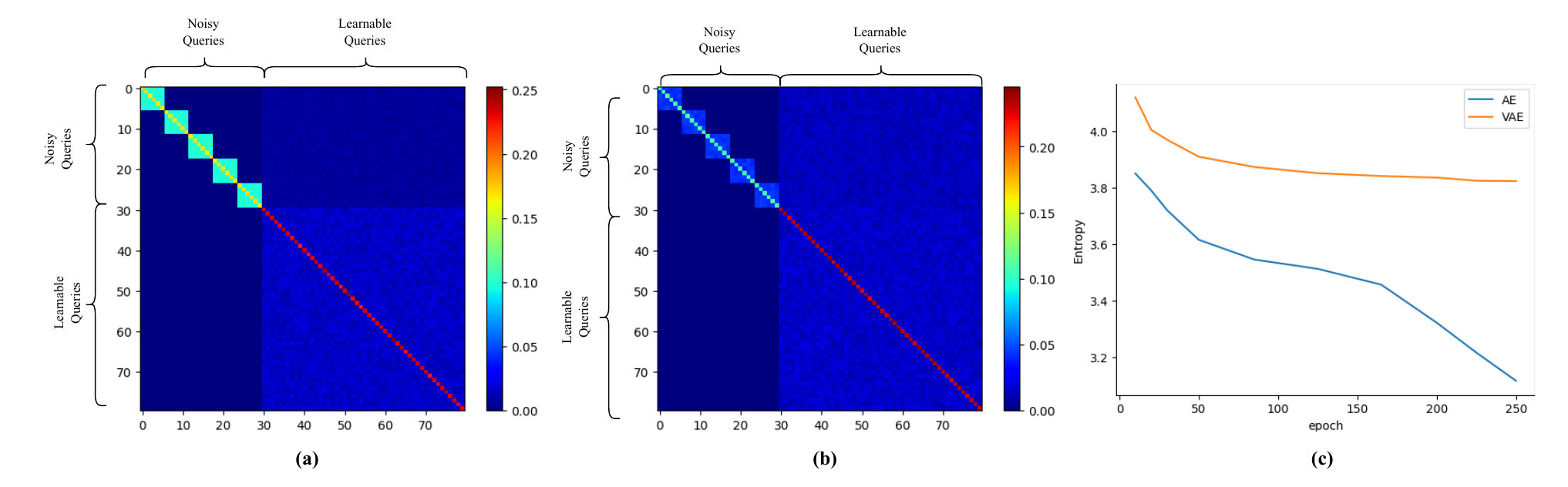}
    \caption{\textbf{The influence of the denoised query on learnable queries.} The self-attention maps trend resulting from the conventional denoising method and the proposed Variational Denoising approach are illustrated in (a) and (b), respectively. (c) presents the average entropy of the attention maps for both methods throughout the entire training period.}
    \label{fig:motivate}
    \vspace*{-\baselineskip}
\end{figure*}
\textbf{Challenges with Conventional Denoising.}
While the integration of conventional denoising techniques offers an initial improvement in model detection performance (Tab.~\ref{tab:contribution_ablation}), this approach encounters the \textbf{gradient vanishing problem}. We can visually demonstrate this issue by inspecting the attention map \(A_i \in \mathbb{R}^{S \times S}\) derived from the Mask Separated Self-Attention mechanism for the \(i\)-th group of learnable queries (Fig.~\ref{fig:motivate}a).
Observation of the attention map after the training process has largely converged, revealing a significant decoupling effect. Specifically, the attention scores between the noisy queries and the learnable queries approach zero, visibly in the upper-right quadrant of the map. This decoupling is detrimental because it obstructs the efficient backpropagation of gradients originating from the reconstruction loss to the learnable queries, thereby limiting their capacity for further adaptation and improvement.

To quantitatively analyze this behavior, we monitored the sparsity of the attention maps \(A_i\) during the evaluation of MonoVQD across training epochs. Treating attention maps as probability distributions, we employed negative entropy as an intuitive metric for sparsity (a lower entropy indicates greater sparsity). Increased sparsity signifies that noisy queries are predominantly attending to themselves rather than the learnable queries. This elevated self-attention among noisy queries directly diminishes the attention weight (and consequently, the gradient flow) directed towards the learnable queries. As illustrated in Fig.~\ref{fig:motivate}c, the attention map associated with the conventional denoising method (AE) exhibits a rapid decrease in negative entropy (increase in sparsity) throughout training, providing further evidence of the restricted gradient flow and resulting performance plateau.

\textbf{Variational Query Denoising.}
To overcome the identified gradient vanishing problem and enhance the effectiveness of the denoising process, we propose leveraging the inherent stochastic properties of a \textbf{Variational Autoencoder (VAE)}. Our approach aims to introduce beneficial variation into the noisy queries, which in turn increases the entropy of the attention distribution and promotes more robust gradient propagation.

As depicted in the overall architecture shown in Fig.~\ref{fig:overall} (Variational Query Generator), the process begins by feeding the initial noisy input boxes through a dedicated box embedding layer. This layer functions as the encoder of our VAE, predicting $\mu$ and $\Sigma$ parameters of a latent distribution. The stochastic noisy queries are then synthesized by sampling from this learned distribution using the reparameterization trick: \(z \sim \mathcal{N}\left(\mu,\Sigma\right)\). These generated stochastic queries are subsequently processed by the model's decoder, analogous to how queries are handled in a standard denoising setup. Training for this denoising process is guided by a denoising loss function, defined as:
\begin{equation}
    \begin{aligned}
        L_{DN} &= L_{res} + \beta L_{KL}\left(\mathcal{N}\left(\mu,\Sigma\right),\mathcal{N}\left(0,I\right)\right)
    \end{aligned}
    \label{eq:denoising}
\end{equation}
Here, $L_{res}$ represents the reconstruction loss computed from the noisy queries, $L_{KL}$ denotes the Kullback-Leibler divergence loss that regularizes the learned distribution towards a standard normal prior $\mathcal{N}(0,I)$, and $\beta$ is a weighting factor.

Empirical results validate our approach. Fig.~\ref{fig:motivate}c demonstrates that the proposed Variational Query Denoising (VAE) maintains significantly higher attention map entropy compared to the conventional denoising method, indicating reduced sparsity. Furthermore, as shown in Fig.~\ref{fig:motivate}b (contrast with Fig.~\ref{fig:motivate}a), the attention scores in the upper-right corner of the self-attention map do not converge to zero. This lack of decoupling confirms that the learnable queries continue to effectively interact with and benefit from the denoising process enabled by our variational approach.
\subsection{Forward-Looking Distillation}
\label{subsec:fld}
We suggest that the self-distillation process can improve the impact of the iterative refinement strategy in DETR~\cite{DETR}. This is achieved by distilling knowledge from the high-performing final decoder layer into shallower layers. We utilize the query set from the final decoder layer to accomplish this. First, we apply the Hungarian Matching to the final layer's predictions to determine the optimal query-to-ground-truth assignments. These matching indices are then used to access the corresponding queries in the earlier decoder layers. By aligning the early decoder's output queries with those of the final layer, we enhance the refinement process as subsequent layers refine their predictions based on increasingly refined inputs. Although the final decoder layer generally produces high-quality queries, some may still exhibit suboptimal performance. To address this, we introduce a weighting mechanism based on the 3D Intersection over the Union ($\mathbf{IoU}_{3D}$). Specifically, we calculate the $\mathbf{IoU}_{3D}$ between the prediction of the last decoder layer and its corresponding ground-truth 3D box. This value $\mathbf{IoU}_{3D}$ is then used to weigh the distillation loss, thereby emphasizing the transfer of knowledge from high-quality predictions.

Inspired by the success of self-distillation~\cite{SD}, we incorporate a shared MLP $\left(f_Q\left(\cdot\right)\right)$ to refine the query before applying the distillation loss. This refinement step improves the quality of the distilled knowledge. The self-distillation loss can be formally expressed as:
\begin{equation}
\begin{aligned}
L_{distill} &= \sum_{i=1}^{L-1}{\mathbf{IoU}_{3D}}_{Q_L}\cdot S_{L1}\left(f_Q\left(Q_i\right), Q_L  \right),
\end{aligned}
\label{eq:loss_distill}
\end{equation}

where $Q_i$ represents the output query of the i-th decoder layer, $L$ denotes the total number of decoder layers in the model, $S_{L1}$ represents the smooth-L1 loss function, $f_Q(Q_i)$ is the fine-tuned version of $Q_i$ obtained from an MLP, ${\mathbf{IoU}_{3D}}_{Q_L}$ is the $\mathbf{Iou}_{3D}$ between the $Q_L$ query and it corresponding 3D ground truth. It is important to note that Forward-Looking Distillation is applied to both the learnable queries $Q_L$ and the noisy queries $Q_N$.
\subsection{Loss Function}
The training loss of MonoVQD is composed of three distinct terms: detection loss $L_{det}$, denoising loss $L_{DN}$, and self-distillation loss $L_{distill}$. We adopt $L_{det}$ formulation from MonoDETR~\cite{MonoDETR} that includes losses for object category, projected center point, 2D bounding box, orientation, 3D size, central depth and depth map. In addition, we propose the denoising loss, $L_{DN}$ derived from Variational Query Denoising as presented in Eq.~\ref{eq:denoising}, and $L_{distill}$ is integrated for the proposed Forward-Looking distillation as mentioned in Eq.~\ref{eq:loss_distill}. The overall loss of MonoVQD is then formulated as:
\begin{equation}
    L_{overall} = \lambda_1L_{det}+ \lambda_{2}L_{DN}+ \lambda_{3}L_{distill}
\end{equation}
\subsection{Plug-and-play for Multi-view Detectors}
\label{subsec:plug_and_play}
Beyond the monocular domain, our Variational Query Denoising and Forward-Looking Distillation demonstrate plug-and-play generalizability. We showcase this by integrating key components into the multi-view RayDN~\cite{RayDN} architecture, preserving its original training configuration. As presented in Tab.~\ref{tab:nuscence_result}, this integration enhances RayDN's denoising capability for hard negative samples and improves its iterative refinement process, thereby leading to improved detection performance.
\begin{table*}[ht]
    \centering
    \resizebox{\textwidth}{!}{
        \begin{tabular}{l|c|ccc|ccc|ccc}
            \toprule
            \multicolumn{1}{l|}{\multirow{2}{*}{Method}} &
            \multicolumn{1}{c|}{\multirow{2}{*}{\hspace{0.05cm} Venue \hspace{0.05cm}}} & \multicolumn{3}{c|}{Test, $AP_{3D}$} &  \multicolumn{3}{c|}{Test, $AP_{BEV}$} &
            \multicolumn{3}{c}{Val, $AP_{3D}$}\\
            & & Easy  & Mod. & Hard & Easy & Mod.  & Hard   & Easy & Mod.  & Hard   \\ \midrule \midrule
            MonoDTR~\cite{MonoDTR} & CVPR'22 & 21.99 & 15.39 & 12.73 & 28.59 & 20.38 & 17.14 & 24.52 & 18.67 & 15.51 \\
            DCD~\cite{DCD} & ECCV'22 & 23.81 & 15.90 & 13.21 & 32.55 & 21.50 & 18.25 & 23.94 & 17.38 & 15.32\\
            MonoJSG~\cite{MonoJSG} & CVPR'22 & 24.69 & 16.14 & 13.64 & 32.59 & 21.26 & 18.18 & 26.40 & 18.30 & 15.40 \\
            MonoCon~\cite{MonoCon} & AAAI'22 & 22.50 & 16.46 & 13.95 & 31.12 & 22.10 & 18.20 & 26.33 & 19.01 & 15.98 \\
            MonoDETR~\cite{MonoDETR} & ICCV'23 & 25.00 & 16.47 & 13.58 & 33.60 & 22.11 & 18.60 & 28.84 & 20.61 & 16.38\\
            MonoCD~\cite{MonoCD} & CVPR'24 & 25.53 & 16.59 & 14.53 & 33.41 & 22.81 & 19.57 & 26.45 & 19.37 & 16.38 \\
            MonoUNI~\cite{MonoUNI} & NeurIPS'23 & 24.75 & 16.73 & 13.49 & 33.28 & 23.05 & 19.39 & 24.51 & 17.18 & 14.01 \\
            OccupanyM3D~\cite{OccupanyM3D} & CVPR'24 & 25.55 & 17.02 & 14.79 & 35.58 & 24.18 & 21.37 & 26.87 & 19.96 & 17.15\\
            OPA-3D~\cite{OPA-3D} & RAL'23 & 24.60 & 17.05 & 14.25 & 33.54 & 22.53 & 19.22 & 24.97 & 19.40 & 16.59 \\
            FD3D~\cite{FD3D} & AAAI'24 & 25.38 & 17.12 & 14.50 & 34.20 & 23.72 & 20.76 & 28.22 & 20.23 & 17.04 \\
            MonoNERD~\cite{MonoNeRD} & ICCV'23 & 22.75 & 17.13 & 15.63 & 31.13 & 23.46 & 20.97 & 20.64 & 15.44 & 13.99\\
            MonoDDE~\cite{MonoDDE} & CVPR'22 & 24.93 & 17.14 & 15.10 & 33.58 & 23.46 & 20.37 & 26.66 & 19.75 & 16.72 \\
            MonoPGC~\cite{MonoPGC} & ICRA'23 & 24.68 & 17.17 & 14.14 & 32.50 & 23.14 & 20.30 & 25.67 & 18.63 & 15.65 \\
            MonoMAE~\cite{MonoMAE} & NeurIPS'24 & \textbf{\textcolor{blue}{25.60}} & \textbf{\textcolor{blue}{18.84}} & \textbf{\textcolor{red}{16.78}} & \textbf{\textcolor{blue}{34.15}} & \textbf{\textcolor{red}{24.93}} & \textbf{\textcolor{red}{21.76}} & \textbf{\textcolor{blue}{30.29}} & \textbf{\textcolor{blue}{20.90}} & \textbf{\textcolor{blue}{17.61}}\\ \midrule
            \textbf{MonoVDQ (ours)} & & \textbf{\textcolor{red}{28.26}} & \textbf{\textcolor{red}{19.20}} & \textbf{\textcolor{blue}{16.21}} & \textbf{\textcolor{red}{35.77}} & \textbf{\textcolor{blue}{24.82}} & \textbf{\textcolor{blue}{21.37}} & \textbf{\textcolor{red}{32.12}} & \textbf{\textcolor{red}{23.55}} & \textbf{\textcolor{red}{20.15}} \\
            \bottomrule
        \end{tabular}
    }
    \caption{\textbf{Detection performance of Car category on the KITTI 3D dataset.} \textbf{\textcolor{red}{Red}} numbers indicates the best results for specific metrics, while \textbf{\textcolor{blue}{blue}} denotes the second-best ones.}
    \label{tab:main_result}
    \vspace*{-\baselineskip}
\end{table*}
\section{Experiments}
\subsection{Settings}
\label{sec:setting}
\textbf{Dataset.} Our model was evaluated on the widely-used KITTI 3D object detection benchmark~\cite{KITTIDATA}. This dataset contains 7481 training images and 7581 testing images. Following the methodology of Chen et al.~\cite{Chen20153DOP}, we split the training set into two subsets: a training set of 3712 images and a validation set of 3769 images. This split facilitated ablation studies to assess the effectiveness of different components within our MonoVQD model.

\textbf{Evaluation metrics.} We evaluated the detection performance in three difficulty levels: easy, moderate, and hard. We used two primary metrics: $AP_{3D}$ and $AP_{BEV}$ indicate the accuracy of 3D bounding box predictions and the accuracy of 2D projections of 3D bounding boxes onto a bird's-eye view respectively. Both $AP_{3D}$ and $AP_{BEV}$ were calculated at 40 recall positions~\cite{simonelli2019disentangling}.

\textbf{Implementation Details.} Our network utilizes the ResNet50 architecture~\cite{ResNet} as its backbone and was trained for 250 epochs using the Adam optimizer with a batch size of 8 images on a single NVIDIA 3090 GPU. The learning rate was initialized at 0.0002 and decayed by a factor of 0.5 at epochs 85, 125, 165, and 225. The weights of losses are set as $\{1, 1, 0.5\}$ for $\lambda_1$ to $\lambda_3$. During the inference phase, queries with a category confidence below 0.2 are discarded, eliminating the need for non-maximum suppression (NMS) post-processing. It is important to note that the proposed denoising method and the self-distillation scheme are used exclusively during the training phase. Consequently, our network maintains the same computational cost, memory consumption, and number of parameters during inference as the base MonoDETR model~\cite{MonoDETR}, while achieving superior performance.

\subsection{Quantitative Results}
\textbf{Experiment on the KITTI 3D test set.} As shown in Tab.~\ref{tab:main_result}, we compare our MonoVQD with several state-of-the-art monocular 3D object detection methods on the KITTI test set. Our approach demonstrates superior performance, surpassing the second-best method, MonoMAE~\cite{MonoMAE}, by a significant margin. Specifically, MonoVQD achieves an improvement of $\textbf{+2.66\%}$ and $\textbf{+0.36\%}$ in $AP_{3D}$ under easy and moderate difficulty levels, respectively. Furthermore, it outperforms MonoMAE~\cite{MonoMAE} by $\textbf{+1.62\%}$ in $AP_{BEV}$ under easy difficulty while still yield similar performance for moderate and hard cases, highlighting the effectiveness of the proposed framework in accurately predicting 3D object locations and dimensions from monocular images.

\textbf{Experiment on the KITTI 3D val set.} We also evaluated our approach on the KITTI validation dataset using the $AP_{3D}$ metrics. As presented in Tab.~\ref{tab:main_result}, our MonoVQD method demonstrates superior performance compared to all existing methods. Notably, it surpasses the second-best approach under easy, moderate, and hard difficulty levels by significant margins: $\textbf{+1.83\%}$, $\textbf{+2.65\%}$, and $\textbf{+1.83\%}$, respectively. These results further emphasize the effectiveness of our proposed framework.

\subsection{Ablation Study}
\textbf{Effectiveness of each proposed component.} In Tab.~\ref{tab:contribution_ablation}, we conduct an ablation study to analyze the effectiveness of
the proposed components: (a) The baseline, MonoDETR~\cite{MonoDETR}, without proposed query enhancement modules; (b) Integrates Forward-Looking Distillation to enhance the effectiveness of iterative refinement strategy; (c) Incorporate vanilla denoising process into the training phase through Mask Seperated Self-Attention to stabilize the Hungarian matching; (d) Enhance denoising effect on the set of learnable queries through Variational Query Denoising.

Firstly, Forward-Looking Distillation (b) significantly improves detection quality, particularly in challenging scenarios (moderate and hard settings), achieving gains of $\mathbf{+0.94\%,+0.87\%}$ in $AP_{3D}$, respectively. This validates the efficacy of our proposed self-distillation approach. Secondly, by incorporating denoising queries and employing a group-wise assignment training scheme through Mask Separated Self-Attention (c), we further enhance model performance. This approach surpasses (b) with gains of $\mathbf{+0.41\%,+1.24\%,+1.18\%}$ in $AP_{3D}$ across easy, moderate, and hard difficulty levels, respectively. This result highlights the ability of Mask Separated Self-Attention to effectively leverage both training schemes concurrently. Finally, integrating Variational Query Denoising (d) yields substantial improvements across all difficulty levels compared to (c), achieving gains of $\mathbf{+1.66\%,+0.77\%,+0.66\%}$, respectively. This significant improvement underscores the effectiveness of incorporating stochastic properties into the denoising process.
\begin{table}[ht]
    \centering
    \begin{tabular}{c|ccc|ccc}
         \toprule
         \multicolumn{1}{l|}{\multirow{2}{*}{}} &
         \multicolumn{3}{c|}{Ablation} & \multicolumn{3}{c}{$AP_{3D}$@IoU=0.7} \\
         & FLD & DN & VDN & Easy & Mod. & Hard \\ \midrule \midrule
         (a) & & & & 29.99 & 20.92 & 17.44 \\
         (b) & \checkmark & & & 30.05 & 21.54 & 18.31 \\
         (c) & \checkmark & \checkmark & & 30.46 & 22.78 & 19.49\\
         (d) & \checkmark & \checkmark & \checkmark & 32.12 & 23.55 & 20.15\\
         \bottomrule
    \end{tabular}
    \caption{\textbf{Analysis of different components of our approach} on the Car category of the KITTI validation set. `FLD' denotes the proposed Forward-Looking Distillation. `DN' denotes 3D query denoising, and `VDN' denotes the proposed variational query denoising.}
    \label{tab:contribution_ablation}
    \vspace*{-\baselineskip}
\end{table}

\textbf{Forward-Looking Distillation.} We analyze the efficacy of each design component of Forward-Looking distillation in Tab.~\ref{tab:self_distillation}. Firstly, incorporating $\mathbf{IoU}_{3D}$ weighting mechanism to ensure quality distillation knowledge from the good query, we can see that the model achieves marked performance. Moreover, using an MLP to refine the early decoder query also improves the detection quality.
\begin{table}[ht]
    \centering
    \begin{tabular}{c|ccc}
         \toprule
         Ablation & \multicolumn{3}{c}{$AP_{3D}$@IoU=0.7} \\
          & Easy & Mod. & Hard \\ \midrule \midrule
         w/o &  29.25 & 21.12 & 17.65 \\
         $\mathbf{IoU}_{3D}$ & 29.71 & 21.32 & 18.05\\
         $\mathbf{IoU}_{3D}$ + MLP & 30.05 & 21.54 & 18.31 \\
         \bottomrule
    \end{tabular}
    \caption{\textbf{The design of Forward-Looking distillation.} w/o denotes directly aligned early query with the last one.`$\mathbf{IoU}_{3D}$' denotes using $\mathbf{IoU}_{3D}$ for weighting between queries. `MLP' denotes using a two-layer MLP to refine the student query.} 
    \label{tab:self_distillation}
    \vspace*{-\baselineskip}
\end{table}
\subsection{Qualitative Results}
\begin{figure*}[t!]
    \centering
    \includegraphics[width=\linewidth]{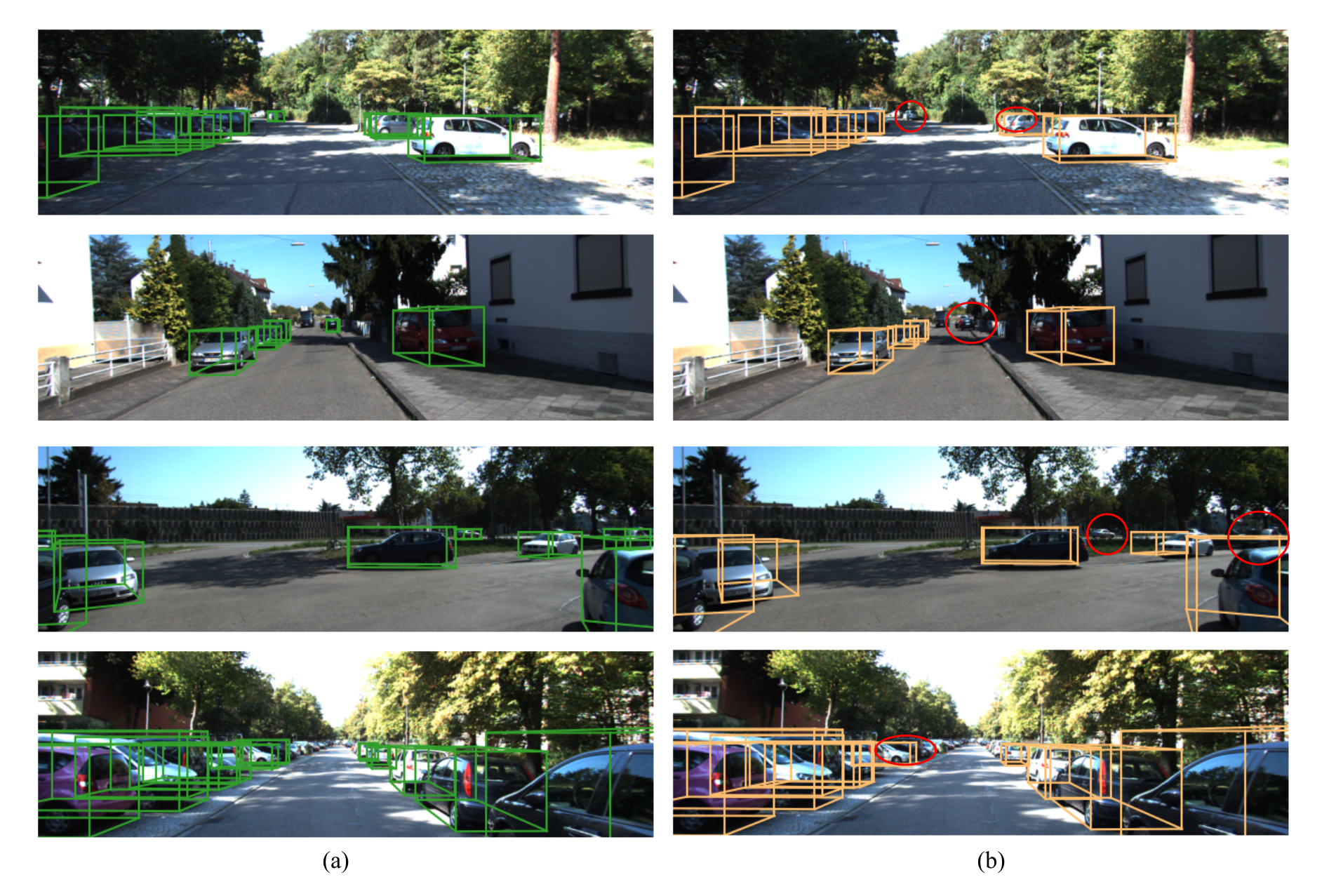}
    \caption{\textbf{Qualitative results on KITTI3D validation set.} (a) The detection results when the model is trained using the proposed Variational Query Denoising. (b) The detection results were obtained when training with a conventional denoising technique. We use \textbf{\textcolor{red}{red}} circle to annotate the missing car in the visualization results.}
    \label{fig:qualitative}
    \vspace*{-\baselineskip}
\end{figure*}

Fig.~\ref{fig:qualitative} presents qualitative results obtained with our method, demonstrating the impact of the proposed Variational Query Denoising technique. As shown, the results with Variational Query Denoising (Fig.~\ref{fig:qualitative}a) exhibit more accurate object localization compared to those without (Fig.~\ref{fig:qualitative}b). For instance, in the third row, the green bounding boxes more accurately enclose the object than the orange bounding boxes. Furthermore, Fig.~\ref{fig:qualitative} shows improved detection of smaller and heavily occluded objects when using Variational Query Denoising. These qualitative improvements corroborate the excellent performance achieved on the KITTI benchmark.
\subsection{Plug-and-play Experiment}
\begin{table*}[ht]    
  \centering
  \resizebox{\textwidth}{!}{%
\begin{tabular}{l | c | c | cc | cccccc}
\toprule
\textbf{Methods}     & \textbf{Backbone}      & \textbf{Image Size}  & \textbf{mAP}  &\textbf{NDS}      & \textbf{mATE}    & \textbf{mASE}          & \textbf{mAOE}       & \textbf{mAVE}     & \textbf{mAAE} \\
\midrule
PETRv2~\cite{liu2023petrv2}      & ResNet50      & 256$\times$704                            & 34.9                   & 45.6                   & 0.700                      & 0.275                    & 0.580                     & 0.437                    & 0.187                    \\
BEVDepth~\cite{li2023bevdepth}    & ResNet50      & 256$\times$704                            & 35.1                   & 47.5                   & 0.629                    & 0.267                    & 0.479                    & 0.428                    & 0.198                    \\
SOLOFusion~\cite{park2022time}  & ResNet50      & 256$\times$704       & 42.7                   & 53.4                   & 0.567                   & 0.274                    & 0.511                    & 0.252                    & 0.181                   \\
SparseBEV~\cite{liu2023sparsebev}$\dagger$   & ResNet50      & 256$\times$704                             & 44.8                   & 55.8                   & 0.581                    & 0.271                    & 0.373                  & 0.247                   & 0.190                     \\
StreamPETR~\cite{Wang_2023_ICCV}$\dagger$  & ResNet50      & 256$\times$704                           & 45.0                    & 55.0                    & 0.613                    & 0.267                    & 0.413                    & 0.265                    & 0.198                    \\
RayDN$\dagger$~\cite{RayDN}  & ResNet50      & 256$\times$704   &46.9 & 56.3 & 0.579 & 0.264 & 0.433 & 0.256 & 0.187 \\
RayDN$\dagger$ (with \textbf{Ours}) & ResNet50 & 256$\times$704 & \textbf{47.8} & \textbf{56.9} & 0.571 & 0.262 & 0.391 & 0.273 & 0.180 \\
\bottomrule
\end{tabular}}
\caption{\textbf{Comparison on the nuScenes validation set.} $\dagger$ Indicates methods that benefit from perspective-view pre-training.}
\label{tab:nuscence_result}
\end{table*}
\textbf{Results on the nuScenes Validation set.} Integrating our module into RayDN~\cite{RayDN}, as clearly demonstrated in Tab.~\ref{tab:nuscence_result}, yields substantial performance gains, evidenced by a noteworthy increase of \textbf{+0.9\%} in mAP and a compelling \textbf{+0.6\%} improvement in NDS. These significant enhancements underscore the effectiveness of our novel training scheme in optimizing the detection capabilities of established architectures. Furthermore, the consistent positive impact on key performance metrics strongly suggests the robust generalization capabilities of our approach, highlighting its potential as a versatile plug-and-play method for readily enhancing a wide range of existing 3D object detection frameworks. This ease of integration and the observed performance boost solidify the practical utility and broad applicability of our proposed module.

\section{Conclusion}
\label{sec:conclusion}
This paper presents MonoVQD, a novel framework designed for the optimization of learnable queries in monocular 3D object detection. MonoVQD incorporates a denoising training regimen with group-wise one-to-many assignments facilitated by Masked-Separate Self-Attention. Through an analysis of attention maps, we identify inherent limitations in conventional denoising methodologies and, consequently, propose Variational Query Denoising to mitigate these deficiencies. Furthermore, Forward-Looking Distillation is introduced to enhance the performance of initial decoder layers, thereby refining the iterative refinement process characteristic of DETR-based architectures. Comprehensive experimental evaluations demonstrate that MonoVQD achieves state-of-the-art performance on established monocular 3D object detection benchmarks. Additionally, a plug-and-play experiment conducted on a multi-view detector showcases the generalization capabilities of the proposed method.

\textbf{Limitations.} The proposed Variational Query Denoising and Forward-Looking Distillation techniques are generally applicable to any DETR-based model; however, this work focuses specifically on their application to 3D object detection. Future research will investigate the broader applicability and generalization of these methods to other domains. To further enhance the denoising process, exploring alternative generative models beyond variational autoencoders, such as flow-based and diffusion models, represents a promising direction for future work.

\newpage
\bibliographystyle{plain}
\bibliography{neurips_2025}
\newpage

\end{document}